 \documentclass[pmlr,twocolumn,10pt]{jmlr} 

\usepackage{booktabs}
\usepackage{siunitx}
\usepackage{cleveref}

\usepackage[switch]{lineno}
\usepackage{subcaption}
\usepackage{listings}
\crefname{lstlisting}{Listing}{Listings}
\Crefname{lstlisting}{Listing}{Listings}
\theorembodyfont{\upshape}
\theoremheaderfont{\scshape}
\theorempostheader{:}
\theoremsep{\newline}

\jmlrvolume{297}
\jmlryear{2025}
\jmlrworkshop{Machine Learning for Health (ML4H) 2025} 

 \title[APRIL]{APRIL: Annotations for Policy evaluation with Reliable Inference from LLMs}

\author{%
\Name{Aishwarya Mandyam} \Email{am2@stanford.edu}\\
\addr Stanford University
\AND
\Name{Kalyani Limaye} \Email{limayk@stanford.edu}\\
\addr Stanford University
\AND
\Name{Barbara E. Engelhardt*} \Email{barbarae@stanford.edu}\\
\addr Stanford University, The Gladstone Institutes
\AND
\Name{Emily Alsentzer*} \Email{ealsentzer@stanford.edu}\\
\addr Stanford University
}

\begin{document}
\maketitle

\begin{abstract}
Off-policy evaluation (OPE) estimates the value of a contextual bandit policy prior to deployment. As such, OPE plays a critical role in ensuring safety in high-stakes domains such as healthcare. However, standard OPE approaches are limited by the size and coverage of the behavior dataset. 
While previous work has explored using expert-labeled counterfactual annotations to enhance dataset coverage, obtaining such annotations is expensive, limiting the scalability of prior approaches.  
We propose leveraging large language models (LLMs) to generate counterfactual annotations for OPE in medical domains. Our method uses domain knowledge to guide LLMs in predicting how key clinical features evolve under alternate treatments. These predicted features can then be transformed using known reward functions to create counterfactual annotations.
We first evaluate the ability of several LLMs to predict clinical features across two patient subsets in MIMIC-IV, finding that state-of-the-art LLMs achieve comparable performance. Building on this capacity to predict clinical features, we generate LLM-based counterfactual annotations and incorporate them into an OPE estimator. Our empirical results analyze the benefits of counterfactual annotations under varying degrees of shift between the behavior and target policies. We find that in most cases, the LLM-based counterfactual annotations significantly improve OPE estimates up to a point. We provide an entropy-based metric to identify when additional annotations cease to be useful. 
Our results demonstrate that LLM-based counterfactual annotations offer a scalable approach for addressing coverage limitations in healthcare datasets, enabling safer deployment of decision-making policies in clinical settings. 
\end{abstract}
\paragraph*{Data and Code Availability}
This paper uses data from the broadly available MIMIC-IV dataset. Code is available on 
\href{https://github.com/aishwarya-rm/april.git}{Github}. 
\paragraph*{Institutional Review Board (IRB)}
This research does not require IRB approval.

\section{Introduction}
Off-policy evaluation (OPE) methods estimate the value of a new (target) contextual bandit policy using a behavior dataset of samples collected under a distinct behavior policy~\citep{sutton2018reinforcement}. OPE can be particularly useful in high-stakes domains such as healthcare, where evaluating policies by directly deploying them is either impossible or unethical. Standard approaches to OPE include importance sampling~\citep{precup_ope}, the direct method~\citep{beygelzimer2009direct}, and doubly robust approaches~\citep{dudik2014doubly}. However, the performance of OPE estimators is inherently limited by the coverage of the behavior dataset. When the target policy takes actions that are under-observed in the behavior dataset, standard OPE methods cannot reliably estimate the value of these actions, leading to inaccurate policy value estimates.

To address this, recent work proposes augmenting the behavior dataset with counterfactual annotations~\citep{tang2023counterfactualaugmented}. A counterfactual annotation is a prediction of the scalar reward resulting from an action unobserved in the behavior dataset. For example, if a patient received $20$mEq of potassium, a counterfactual annotation would predict the reward had the patient instead received $40$mEq. Two strategies have been developed to incorporate such annotations into OPE: one augments an importance sampling–based estimator~\citep{tang2023counterfactualaugmented}, and the other augments a doubly robust estimator~\citep{mandyam2024candorcounterfactualannotateddoubly}. Both demonstrate that incorporating counterfactual annotations can improve OPE estimates, but these approaches rely on human experts (e.g., clinicians) to provide the annotations, which is costly and difficult to scale.

To address this, we propose a pipeline to source counterfactual annotations for OPE in clinical settings using large language models (LLMs). LLMs have the ability to reason effectively about medical domains, with the capacity to answer medical questions~\citep{singhal2023expertlevelmedicalquestionanswering}, perform differential patient diagnoses~\citep{nori2025sequentialdiagnosislanguagemodels}, and reason about medical images~\citep{zhou2025medversageneralistfoundationmodel}. Our approach leverages LLMs to predict clinical features of interest such as downstream laboratory measurements; we then incorporate these predictions into known reward functions to produce synthetically generated counterfactual annotations. 

We evaluate our proposed framework on two clinical tasks: intravenous (IV) potassium and sodium repletion. Both are critical procedures in clinical practice, where large errors in administration can lead to adverse outcomes~\citep{voldby_brandstrup}. Furthermore, these are routine procedures with well-established guidelines for treatment and reasonably predictable treatment response curves, making them especially tractable settings for applying contextual bandit algorithms. We construct corresponding patient datasets from the Medical Information Mart for Intensive Care IV (MIMIC-IV) database, which contains electronic health records (EHR) for patients admitted to the Beth Israel Deaconess Medical Center~\citep{johnson2024mimiciv,johnson2023mimicivpaper,goldberger2000physionet}. We first assess the ability of several LLMs to predict relevant clinical features, including serum potassium and sodium values. Using clinically motivated reward functions, we then transform these predictions into counterfactual annotations. Our results show that LLM-generated counterfactual annotations improve OPE estimates, particularly under large distribution shifts between the behavior and target policies.

Our contributions follow:
\begin{itemize}
    \item \textbf{We perform OPE with LLM-generated counterfactual annotations in a multi-cohort setting} using MIMIC-IV. We systematically evaluate multiple general-purpose LLMs for their accuracy in predicting downstream clinical features.
    \item \textbf{We show that incorporating LLM-generated annotations can significantly improve OPE estimates}, reducing RMSE relative to baselines and confirming prior findings in real-world data.
    \item \textbf{We demonstrate that additional counterfactual annotations offer diminishing returns}, a phenomenon captured quantitatively via the marginal entropy over the action distribution.
\end{itemize}

\section{Preliminaries}
\subsection{Problem setting}
We adopt a contextual bandit setting, as potassium and sodium repletion are short-horizon decisions whose outcomes can be observed within a single timestep. A contextual bandit setting is represented as 
$(\mathcal{S}, \mathcal{A}, \mathcal{R}, d_0)$, where $\mathcal{S}$ is the discrete context space, $\mathcal{A}$ is the discrete action space, $\mathcal{R}$ is the reward distribution, and $d_0$ is the initial context distribution. The reward function $R: S \times A \to [0, 1]$ assigns a scalar reward between $0$ and $1$. Our goal is to evaluate a target contextual bandit policy $\pi_e$ by estimating its value $v(\pi_e) = \mathbb{E}_{s \sim d_0, a \sim \pi_e,)}\left[R(s, a)\right]$ using a behavior dataset. The behavior dataset consists of samples $D=\{s_i, a_i, r_i\}_{i=1}^N$, where the actions are sampled from a behavior policy \(\pi_b\).

\subsection{Off-policy evaluation estimators}
Many OPE estimators fall into three broad categories: importance sampling (IS), the direct method (DM), and doubly robust (DR) estimators. IS methods~\citep{precup_ope} re-weigh each sample in the behavior dataset using an inverse propensity score (IPS) $\frac{\pi_e(a_i|s_i)}{\pi_b(a_i|s_i)}$. The second class includes direct-method (DM) approaches~\citep{beygelzimer2009direct}, which learn a reward model $\hat{R}$ from the behavior dataset, and use the model to simulate the returns of samples from the target policy. The final category includes doubly-robust (DR) approaches~\citep{dudik2014doubly,jiang2016doubly}, which combine strategies from IS and DM approaches, providing favorable theoretical guarantees when either the IPS ratio is known or the reward model is of high quality. 

Recent work has proposed supplementing the behavior dataset with counterfactual annotations solicited from an expert. \citet{tang2023counterfactualaugmented} introduce an IS-based estimator and demonstrate that counterfactual annotations can improve OPE estimates when the annotations are of high quality. \citet{mandyam2024candorcounterfactualannotateddoubly} extends this to a doubly robust setting, mitigating the negative impacts of noisy or imperfect annotations. Both approaches assume that counterfactual annotations are expert-labeled, which limits the scalability of the proposed approaches. Other work has proposed using a variational auto-encoder to generate synthetic trajectories, thus enriching state–action coverage of the behavior dataset and tightening variance bounds~\citep{gao2024trajectory}. Our work builds on these approaches, identifying a scalable alternative to creating counterfactual annotations.

\subsection{Generative models can encode medical knowledge}
LLMs have shown impressive general medical reasoning capabilities. Models fine‑tuned on web and biomedical corpora now match or surpass physicians on multiple‑choice benchmarks such as MedQA~\citep{jin2020medqa}. DeepMind’s Med-PaLM 2~\citep{singhal2023medpalm} and Gemini models~\citep{saab2024capabilitiesgeminimodelsmedicine} illustrate that scaling and instruction tuning can boost performance across a range of clinical knowledge tasks. However, these works center on general medical knowledge questions rather than reasoning about individual patient clinical trajectories, which is the focus of our work.

More granular, patient-specific LLM applications are beginning to emerge, including reasoning about how laboratory values evolve over a patient trajectory. ~\citet{bhasuran2025preliminary} explore differential-diagnosis generation from brief clinical vignettes, highlighting the importance of structured patient summaries to improve LLM outputs. ~\citet{he2024quality} evaluate the ability to generate accurate and safe responses to patient lab-result inquiries using prompt engineering and detailed quality evaluation metrics. These studies suggest that LLMs can reason about patients when provided with curated input and task framing~\citep{wei2022chain, chung2022scaling}. Our work leverages prompting strategies that build on those used in these works to guide LLMs in generating patient-specific counterfactual annotations.

\subsection{Synthetic data for machine learning}
In our setting, supervision comes from both a real-world dataset and a noisier set of synthetic data. Using a noisy secondary dataset is a common paradigm in supervised learning, and methods to mitigate the covariance shift between the datasets have been extensively studied in the \emph{robust machine learning} literature. Earlier results introduced transfer learning techniques to learn features from secondary datasets while mitigating issues with higher-variance samples~\citep{Krizhevsky2012ImageNetCW, transfer_learning, domain_adaptation,covariance_shift, covariate_shift_ml}. Other methods such as prediction-powered inference~\citep{angelopoulos2023predictionpoweredinference} explicitly correct for possible biases that result from the introduction of synthetic samples. 

\section{Methods}
\begin{figure}[htbp]
\floatconts
  {fig:pipeline}
  {\caption{\textbf{Our work improves OPE estimates using LLM-generated counterfactual annotations.} We first query counterfactual annotations using domain knowledge guided prediction. We calculate the annotations using a known reward function $R$. Finally, we incorporate the counterfactual annotations and offline behavior dataset to learn an OPE estimate $\hat{v}(\pi_e)$. }}
{\includegraphics[width=\linewidth]{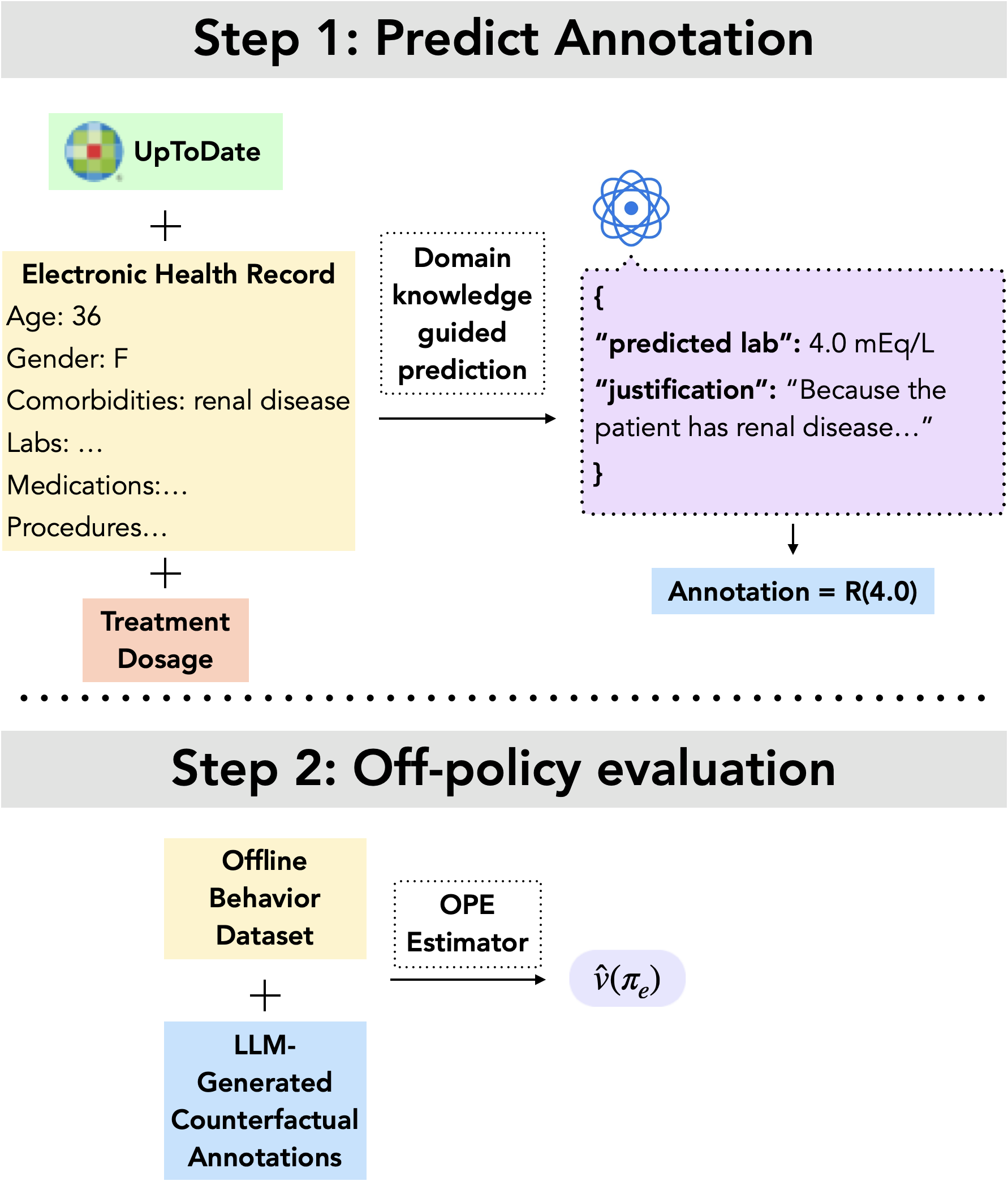}}
\end{figure}

\subsection{Dataset}
\label{sec:dataset}
We conduct our analysis using the MIMIC-IV dataset, partitioned into two subsets of non-ICU patients. The first subset includes all patients who received IV potassium, and the second includes all patients who received IV hypertonic (3\%) saline.  
For each patient in the potassium and sodium subsets, we represent the clinical context as a feature vector comprising $15$ variables that characterize the patient’s state four hours prior to treatment (i.e., administration of potassium or saline). We focus on a four-hour window because this corresponds to the highest frequency of electrolyte administration observed in our dataset, with patients receiving electrolytes at most once every four hours. The features used to represent the clinical context include laboratory results, vital signs, administered medications, and static covariates such as age and gender. A complete list of features is provided in \Cref{apd:dataset}.  The action space corresponds to the administered dosage, represented in milliequivalents (mEq). For potassium administration, the dosage action space is $A = \{0, 10, 20, 40\}$. 
For sodium (i.e., hypertonic saline) administration, dosages are discretized to accommodate our assumption of a discrete action space, yielding an action space $A = \{0, 100, 200, 300, 400, 500\}.$

Similar to prior work~\citep{prasad2020methods}, we adopt a reward function defined as a function of the clinical context observed following the administration of a treatment dosage. Specifically, the scalar reward depends on a single laboratory measurement in the next observed context. For patients who receive IV potassium, this is a serum potassium lab, and for patients who receive IV hypertonic saline, this is a serum sodium lab. The reward function 
\[
R(x) =
\begin{cases} 
\exp\Big(-\frac{1}{2}\left(\frac{x - a}{2.5}\right)^2\Big), & x < a \\[2mm]
1, & a \le x \le b\\[1mm]
\exp\Big(-\frac{1}{2}\left(\frac{x - b}{2.5}\right)^2\Big), & x > b
\end{cases}
\]

takes as input a laboratory value $x$, and uses the lower bound ($a$) and upper bound ($b$) of the reference range to calculate a scalar reward. This reward function design reflects the clinical goal of repletion which is to bring a patient’s electrolyte level into the normal range and keep it there, while smoothly penalizing deviations outside the range. Visual representations of the reward function can be seen in Appendix \Cref{fig:rewards}. 

\subsection{Generating Counterfactual Annotations using LLMs}
\label{sec:prompt_annotations}
As described in \Cref{sec:dataset}, the reward functions for each decision-making task are functions of a single lab value. Therefore, generating a counterfactual annotation requires predicting the specified lab value under a counterfactual treatment dosage.

We construct prompts that include information about the  patient's clinical state, a paragraph that cites the most relevant features for lab value prediction sourced from UpToDate~\citep{uptodate}, and a query about the lab value had an alternative treatment dosage been administered (example in \Cref{apd:prompts}). Building on prior work~\citep{hegselmann2025largelanguagemodelspowerful}, we organize the patient's information into categories such as comorbidities, laboratory results, and medications to create a structured text representation of the clinical state. Including features from UpToDate guides the LLM toward clinically relevant information, as EHRs often contain extraneous data that may not be predictive. To ensure structured outputs, we restrict the LLM's response to a JSON object containing two keys: the predicted lab value, and a justification for the prediction. This format allows for straightforward extraction of the numerical lab value and facilitates verification of the LLM's reasoning.

For potassium administration, dosages are assumed to be delivered at a rate of $10,\mathrm{mEq/hr}$, and for sodium administration, at a rate of $30,\mathrm{mEq/hr}$, corresponding to the most common rates observed in MIMIC-IV. The prompt also specifies that the lab value should be predicted three hours after the IV infusion concludes; this corresponds to the average number of hours that the lab value post treatment administration was measured. Once the LLM predicts the lab value, it is converted into a scalar counterfactual annotation using the corresponding known reward function.

\subsection{Incorporating Counterfactual Annotations into an OPE estimator}
Once we generate counterfactual annotations, we must  incorporate them into an OPE estimator. Prior methods for OPE with counterfactual annotations often assume that the IPS ratios are fully known. However, in this work, we must infer both $\pi_b$ and $\pi_e$ from finite sample sizes. To mitigate possible biases as a result of unknown IPS ratios, we choose to use a direct method estimator. The standard direct method estimator is 
$$\hat{V}^{DM} = \sum_{s \in S} d_0(s) \sum_{a \in A} \pi(a|s) \hat{R}(s,a),$$ 
where $\hat{R}$ is a reward function estimate learned from the behavior dataset. When we have access to both a behavior dataset and counterfactual annotations, we choose to use modified version of the standard DM estimator suggested by prior work work~\citep{mandyam2024candorcounterfactualannotateddoubly}, 
$$\hat{V}^{DM^+} = \sum_{s \in S} d_0(s) \sum_{a \in A} \pi(a|s) \hat{R}^+(s,a),$$
where $\hat{R}^+$ is learned using both the behavior dataset and counterfactual annotations. In this work, we approximate both $\hat{R}$ and $\hat{R}^+$ using linear regression. 

\subsection{Evaluation Setup}
\label{sec:eval_setup}
A standard metric for assessing the accuracy of an OPE estimator is the root mean squared error (RMSE), defined as
\[
\text{RMSE} = \sqrt{\mathbb{E}[(\hat{v}(\pi_e) - v(\pi_e))^2]},
\]
where $\hat{v}(\pi_e)$ denotes the value estimated by the OPE method, and $v(\pi_e)$ is the true value of the target policy $\pi_e$. In practice, $v(\pi_e)$ is rarely available, which complicates the evaluation of OPE estimators in real-world settings.

To address this limitation in the MIMIC-IV dataset, we adopt a controlled evaluation strategy. We partition each dataset subset into disjoint behavior and target sub-cohorts, and infer corresponding policies via behavior cloning. Because the target sub-cohort contains observed rewards, we approximate the value of the cloned target policy by averaging these rewards. The fidelity of this approximation depends on how well the policies are cloned; to assess this, we evaluate the cloned policies' accuracy on a held-out validation set and find that they perform well in reproducing the observed treatment decisions (e.g., validation accuracy > 90\%). This gives us confidence that the averaged rewards in the target subset provide a reliable reference value against which to compute RMSE for different OPE estimators.

It is well known that the performance of OPE estimators depends considerably on the distribution shift between the behavior dataset and the samples induced by the target policy. To systematically study the effect of LLM-generated counterfactual annotations on an OPE method under varying degrees of distribution shift, we construct three behavior–target dataset pairs for each subset of patients from MIMIC-IV. The first pair splits by gender, with female patients forming the behavior dataset and male patients the target dataset. The second pair splits by comorbidity status: for potassium repletion, patients without renal disease form the behavior dataset and patients with renal disease the target dataset; for sodium repletion, the split is based on cirrhosis. We choose these comorbidities because their presence is likely to influence the patient's response to drug administration. The third pair separates patients by drug dosage, using low-dosage patients as the behavior dataset and high-dosage patients as the target dataset. These partitions are designed to reflect clinically meaningful subgroups while also inducing progressively larger divergences between the behavior and target policies. This allows us to evaluate when counterfactual annotations generated by LLMs yield improvements in OPE accuracy.

\section{Experiments}
Our empirical analyses seek to answer the following questions: \textbf{(1)} Can LLMs accurately predict downstream patient laboratory values after a treatment is administered? \textbf{(2)} Under what conditions do LLM-generated counterfactual annotations improve OPE estimates? \textbf{(3)} How do OPE estimates vary as the number of synthetic counterfactual annotations increases?

To address these questions, we use five LLMs spanning a range of parameter counts: OpenAI’s \texttt{o1}~\citep{openai2024openaio1card}, \texttt{o3-mini}~\citep{ZhangOpenAIOS}, and \texttt{gpt-4o-mini}~\citep{Hurst2024GPT4oSC}, Google’s \texttt{Gemini} 1.5~\citep{Reid2024Gemini1U}, and Anthropic’s \texttt{Claude 3.7 Sonnet}~\citep{Claude3S}. All models are hosted on an internal, sandboxed cluster to ensure HIPAA compliance with the MIMIC-IV dataset. All experiments were conducted with a temperature setting of zero whenever supported. For \texttt{o1} and \texttt{o3-mini}, which do not expose a temperature parameter, we use the default configuration.

\subsection{LLMs can predict downstream lab values on real patient populations}
\begin{table*}[ht]
\centering
\begin{tabular}{llccccc}
\hline
\textbf{Task} & \textbf{Cohort} & \textbf{o1} & \textbf{gpt-4o-mini} & \textbf{o3-mini} & \textbf{Gemini} & \textbf{Claude 3.7} \\
\hline
Potassium Repletion & Gender      & 0.856 & 0.809 & 0.854 & 0.858 & \textbf{0.866} \\
                    & Comorbidity & 0.871 & 0.787 & 0.869 & 0.872 & \textbf{0.879} \\
                    & Dosage      & 0.878 & 0.791 & 0.876 & 0.879 & \textbf{0.885} \\
Sodium Repletion    & Gender      & 0.758 & 0.774 & 0.738 & 0.749 & \textbf{0.776} \\
                    & Comorbidity & 0.771 & \textbf{0.801} & 0.753 & 0.779 & 0.796 \\
                    & Dosage      & 0.772 & \textbf{0.809} & 0.768 & 0.780 & 0.804 \\
\hline
\end{tabular}
\caption{\textbf{All LLMs perform comparably across potassium and sodium lab prediction.} Predictions are evaluated using weighted F1 scores across clinically relevant lab value categories. The best performing LLM within each cohort is in bold.}
\label{tab:lab_preds}

\end{table*}

We first evaluate whether LLMs can accurately predict serum potassium and serum sodium laboratory values. In realistic deployment, the target patient population may not be directly accessible, so we assess predictive performance using behavior datasets from each sub-cohort split. To generate predictions, we prompt the LLM following the procedure in \Cref{sec:prompt_annotations}, but instead of asking for counterfactual lab values, we request the lab value following the dosage administered in MIMIC-IV. Because the corresponding ground-truth lab values are observed in MIMIC-IV, we can directly quantify predictive accuracy. We evaluate accuracy using a weighted F1 score across clinically relevant categories of lab values (e.g., below reference range, within reference range, above reference range). The categories used to calculate the F1 score, and further details are reported in \Cref{sec:dataset}. 

We find that LLMs can predict serum potassium and serum sodium lab values with clinically meaningful degrees of accuracy (\Cref{tab:lab_preds}, visualized in Appendix \Cref{fig:lab_preds}). First, we note that serum potassium lab values are predicted more accurately than serum sodium lab values, likely due to the wider distribution and higher prevalence of outliers in sodium lab measurements. We also find that the performance of a given LLM remains consistent across cohorts within each prediction task, which suggests that predictive accuracy does not strongly depend on the underlying patient population. Finally, the differences in predictive accuracy across LLMs are modest, suggesting that multiple models are capable of producing reliable predictions of downstream lab values. Furthermore, these results demonstrate our proposed framework's ability to produce counterfactual annotations of reasonable quality. In particular, because LLMs can predict downstream lab values within a degree of accuracy that is clinically relevant, the resulting annotations are likely to be useful for OPE. 

\subsection{LLM-produced counterfactual annotations can improve OPE estimates}
\begin{figure*}[htbp]
\floatconts
  {fig:ope}
  {\caption{\textbf{LLM-generated counterfactual annotations improve OPE estimates in settings with high divergence between actions observed in behavior and target policies.} We report results for the potassium repletion task. Our baseline is a direct method estimator (blue) that does not use counterfactual annotations. The performance of $DM^+$ with annotations from each LLM is reported in the corresponding colors. Error bars represent standard error across $500$ bootstrapped datasets sampled with replacement. Since RMSE is non-negative, the lower bound of the error bars is truncated at 0 where necessary. Figures above each plot demonstrate the difference in distribution of actions observed in the behavior and target policies.}}
{\includegraphics[width=0.85\linewidth]{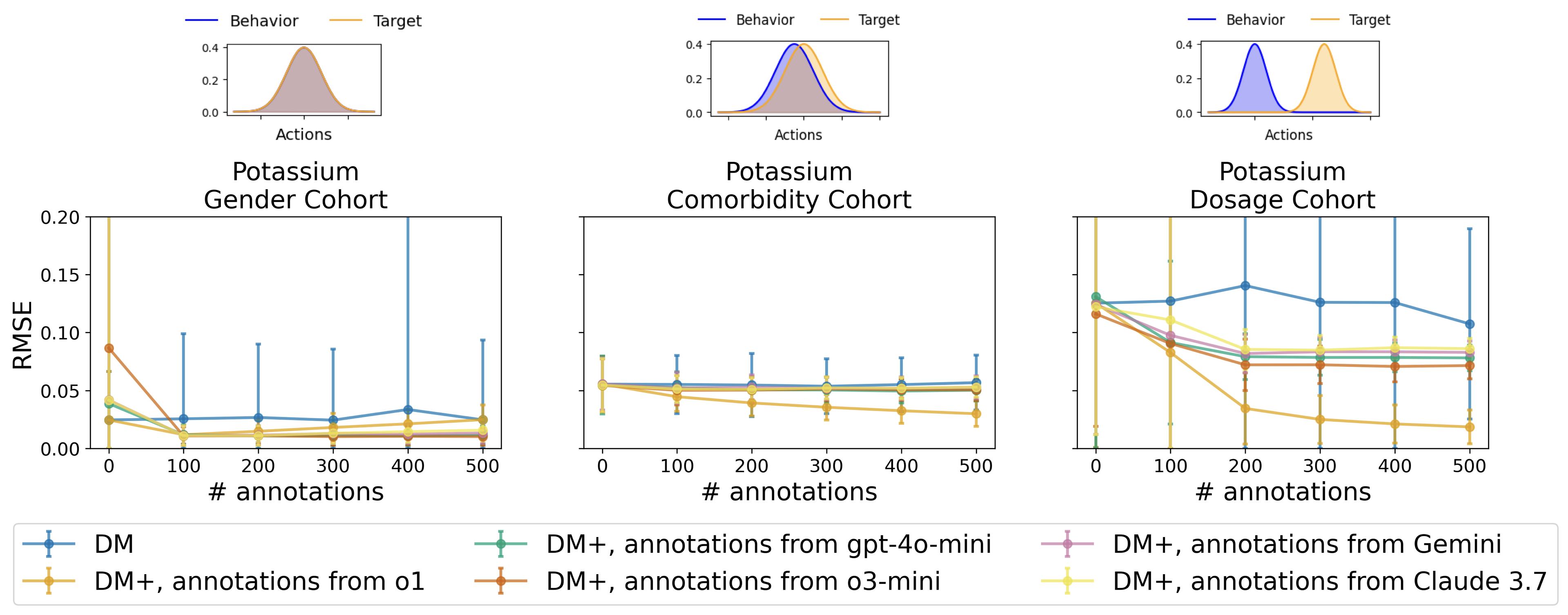}}
\end{figure*}

\begin{table*}[ht]
\centering
\label{tab:p_values}
\begin{tabular}{llccccc}
\hline
\textbf{Task} & \textbf{Cohort} & \textbf{o1} & \textbf{gpt-4o-mini} & \textbf{o3-mini} & \textbf{Gemini} & \textbf{Claude 3.7} \\
\hline
Potassium Repletion & Gender      & \textcolor{red}{5.4E-31} & 9.8E-03 & 3.0E-04 & \textcolor{red}{3.7E-01} & \textcolor{red}{2.4E-01} \\
                    & Comorbidity & 1.5E-94 & 1.3E-08 & 1.1E-07 & 7.7E-03 & 5.7E-04 \\
                    & Dosage      & 1.3E-83 & 1.7E-14 & 1.9E-20 & 7.2E-11 & 1.4E-08 \\
Sodium Repletion    & Gender      & 1.9E-03 & \textcolor{red}{7.0E-04} & 1.5E-19 & 7.5E-14 & 6.6E-08 \\
                    & Comorbidity & 1.0E-07 & 2.0E-16 & 2.0E-03 & 1.0E-04 & 2.2E-05 \\
                    & Dosage      & 2.8E-37 & 3.0E-15 & 3.2E-57 & 1.44E-44 & 7.79E-39 \\
\hline
\end{tabular}
\caption{\textbf{In most cohorts across both tasks, LLM-generated annotations significantly improve RMSE.} We compare RMSE distributions for $DM$ and $DM^+$ with $500$ counterfactual annotations using a paired t-test, and report p-values. P-values shown in red indicate results that are not statistically significant ($p \geq 0.05$) or cases where RMSE does not improve relative to $DM$ ($t < 0$).
}
\end{table*}
We next evaluate the utility of LLM-generated counterfactual annotations for OPE, following the setup in \Cref{sec:eval_setup}. We report results for the potassium repletion task in \Cref{fig:ope}, and for the sodium repletion task in Appendix \Cref{fig:ope_hn}. Our results show that counterfactual annotations substantially improve OPE estimates in settings with large distribution shifts between the actions observed in the behavior and target policies. Across both the potassium and sodium repletion tasks, the reported RMSE reflects the relative difficulty of estimating $v(\pi_e)$ under each cohort split. For example, in the gender cohort split, where behavior and target policies are nearly identical, the RMSE is already near zero without counterfactual annotations, leaving little room for improvement. In contrast, in the dosage cohort split, where behavior and target policies have little overlap, the baseline RMSE of $DM$ is highest, reflecting the difficulty of the task. Here, the incorporation of counterfactual annotations produces the largest reductions in RMSE, indicating that annotations are most valuable when behavior and target policies diverge strongly. Specifically, in the potassium dosage cohort, counterfactual annotations can reduce RMSE by 83\%, and in the sodium dosage cohort by 49\%.

We also find that the performance of $DM^+$ varies with the choice of LLM used to generate counterfactual annotations. In the potassium repletion task, annotations from \texttt{o1} yield the best performance as shown by lowest RMSE, whereas in the sodium repletion task, annotations from \texttt{gpt-4o-mini} and \texttt{o3-mini} yield the best performance. Although the best-performing LLM is not consistent across tasks or cohort splits, counterfactual annotations consistently reduce RMSE in the most challenging settings (e.g., dosage cohorts), regardless of the LLM used.

Finally, to assess statistical significance, we compare $DM$ and $DM^+$ using a paired t-test, with $DM^+$ learned using $500$ counterfactual annotations (\Cref{tab:p_values}). In nearly all settings, $DM^+$ achieves significantly lower RMSE than $DM$. The main exception is the gender split in both potassium and sodium tasks, where annotations from some LLMs do not yield a meaningful performance improvement. This outcome is expected, given the substantial overlap between behavior and target policies in the gender cohorts, which allows $DM$ to perform well even without counterfactual annotations.

\subsection{Additional counterfactual annotations yield marginal improvements in OPE estimates}
\begin{figure}[htbp]
\floatconts
  {fig:practical}
  {\caption{\textbf{Combining annotation sources yields limited returns.} (Top) We compare $DM$ to $DM^+$ with annotations from the best-performing LLMs for potassium repletion in the dosage cohort, with two aggregation methods: pooling predictions and averaging annotations. Error bars show standard error over $500$ bootstrapped datasets, truncated at 0. (Bottom) Marginal entropy over the action space $H(A)$ when adding counterfactual annotations to the behavior dataset for the potassium cohort. The dashed line marks the maximum possible entropy.
  }}
{\includegraphics[width=0.75\linewidth]{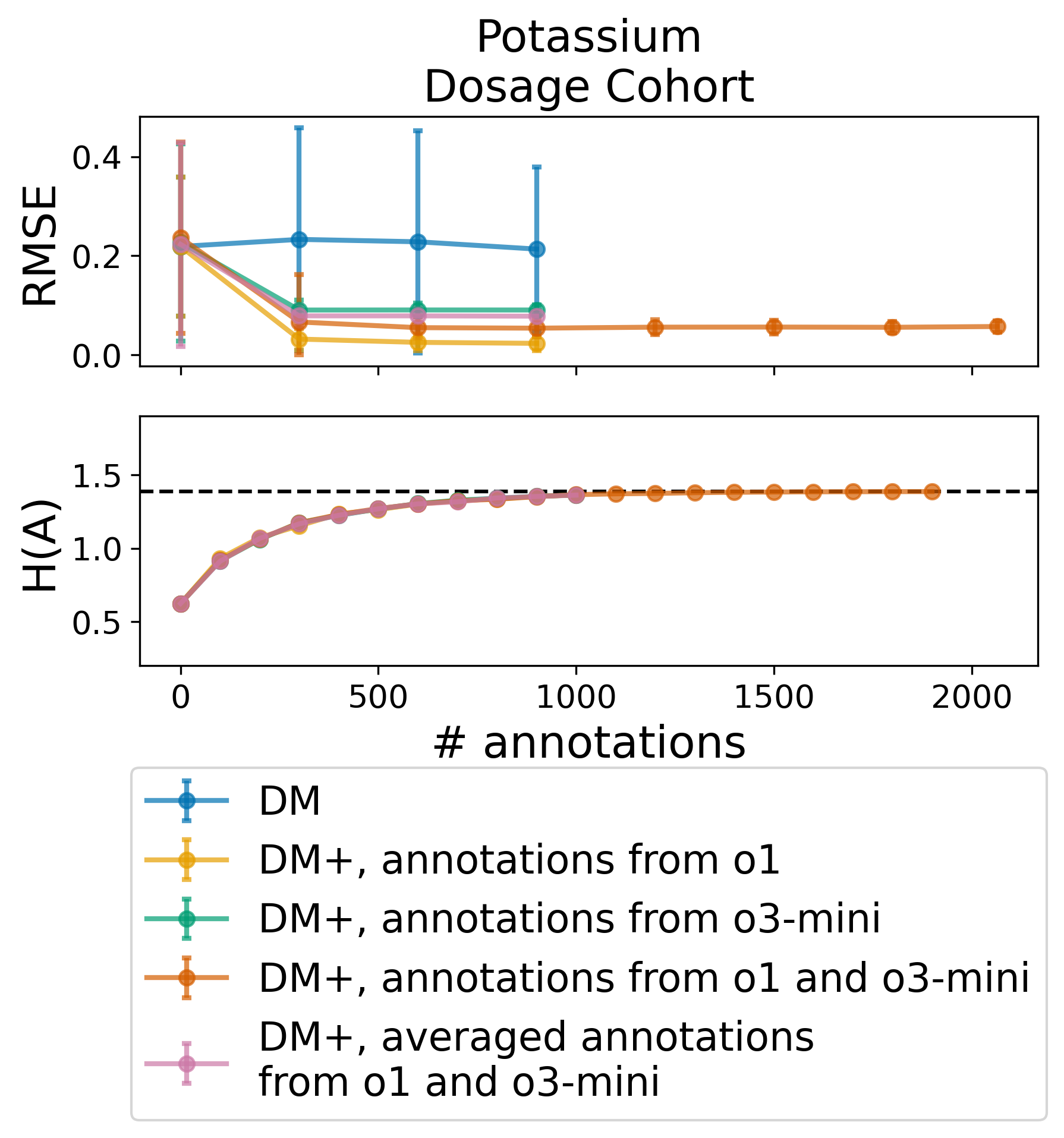}}
\end{figure}
A key consideration when using synthetic data in machine learning is determining the point at which adding further synthetic samples no longer provides benefits. In our setting, a single source of counterfactual annotations can generate at most $N \cdot (|\mathcal{A}|-1)$ unique annotations, where $|\mathcal{A}|$ is the number of actions and $N$ is the number of samples in the behavior dataset. When multiple sources are available, each source provides separate predictions for unobserved actions, which can either be combined or averaged. Direct combination increases the total number of annotations, whereas averaging maintains the same total count. We study both strategies for the potassium task (\Cref{fig:practical}) and sodium task ( \Cref{fig:practical_hn}).

We focus on the dosage cohort splits for both tasks, where counterfactual annotations have the greatest impact in reducing RMSE due to minimal overlap between the behavior and target policies. Specifically, we examine combinations of the two LLMs whose counterfactual annotations yield the best performance for $DM^+$: \texttt{o1} and \texttt{o3-mini} for the potassium task, and \texttt{Gemini} and \texttt{o3-mini} for the sodium task. We find that, while adding counterfactual annotations initially reduces OPE error, the improvement quickly plateaus as more annotations are included. Averaging multiple sources does not provide additional gains beyond the best-performing single source. For instance, in the potassium task, averaging annotations from \texttt{o1} and \texttt{o3-mini} yields OPE performance worse than using \texttt{o1} alone, though slightly better than \texttt{o3-mini} alone. Similarly, combining annotations without averaging, which nearly doubles the number of annotations, does not improve OPE estimates relative to a single source. These results indicate that substantially increasing the number of counterfactual annotations provides limited utility. 

To quantify the effect of additional annotations, we compute the marginal entropy over the action distribution. Entropy measures the overall uncertainty or spread of actions in the dataset. Formally, the marginal entropy over the action distribution is $H(A) = - \sum_{a \in A} \hat{p}(a) \textrm{ln}(\hat{p}(a))$ where $\hat{p}(a)$ is the probability of observing a given action $a$, estimated empirically. The maximum entropy occurs when all actions are equally frequent, in which case $H(A) = ln(|\mathcal{A}|)$. We observe that, as the number of annotations increases, the action coverage approaches the maximum entropy, and further annotations yield only marginal gains. In particular, at around $700$ annotations for the potassium task and $500$ annotations for the sodium task, OPE improvements have largely plateaued, and the marginal action entropy is already near its maximum, indicating that additional counterfactual annotations provide little further utility. This analysis suggests that marginal entropy over the action space is a proxy that may be used to determine when to stop generating counterfactual annotations. 

\section{Discussion}
In this work, we present a scalable strategy for generating counterfactual annotations for OPE in clinical settings. We show that LLMs can reason over clinical contexts and predict downstream lab values, which in turn can be used to construct counterfactual annotations. Focusing on the potassium and sodium repletion tasks, we demonstrate that this approach leads to substantial improvements in OPE estimates, particularly when there is considerable divergence between the behavior and target policies.
We recommend using LLM-generated annotations when there are known coverage gaps in the behavior dataset, and relying on an entropy-based metric to decide when additional counterfactual annotations are needed.

\textbf{Limitations and Future Work.} Our study is limited to reward functions that consider a single clinical feature. While our results provide evidence that LLMs can reliably predict these downstream lab values, future work should evaluate whether similar gains can be achieved for predicting more complex clinical outcomes.

\acks{AM was funded in part by a Stanford Data Science Fellowship. BEE was funded in part by grants from the Parker Institute for Cancer Immunology (PICI), the Chan-Zuckerberg Institute (CZI), the Biswas Family Foundation, NIH NHGRI R01 HG012967, and NIH NHGRI R01 HG013736.
BEE is a CIFAR Fellow in the Multiscale Human Program. The authors would like to thank Dr. Chloe Stanwyck for support with designing and evaluating empirical procedures. }

\bibliography{jmlr-sample}

\newpage
\appendix
\onecolumn

\section{MIMIC-IV dataset}
\label{apd:dataset}
The MIMIC-IV dataset consists of patient data for over 65,000 patients admitted to the ICU and over 200,000 patients admitted to the emergency department. This data is represented as electronic health records (EHRs), which capture a variety of information about each patient including static covariates such as age and gender, all hospital procedures and events such as lab measurements and administered medications, as well as indications of comorbidities. In this work, we consider non-ICU patients who have been administered either IV potassium, or IV hypertonic saline. We have 1622 patients who were administered IV potassium, and 1187 patients who were administered saline. 

\begin{figure}
    \centering
    \includegraphics[width=0.6\linewidth]{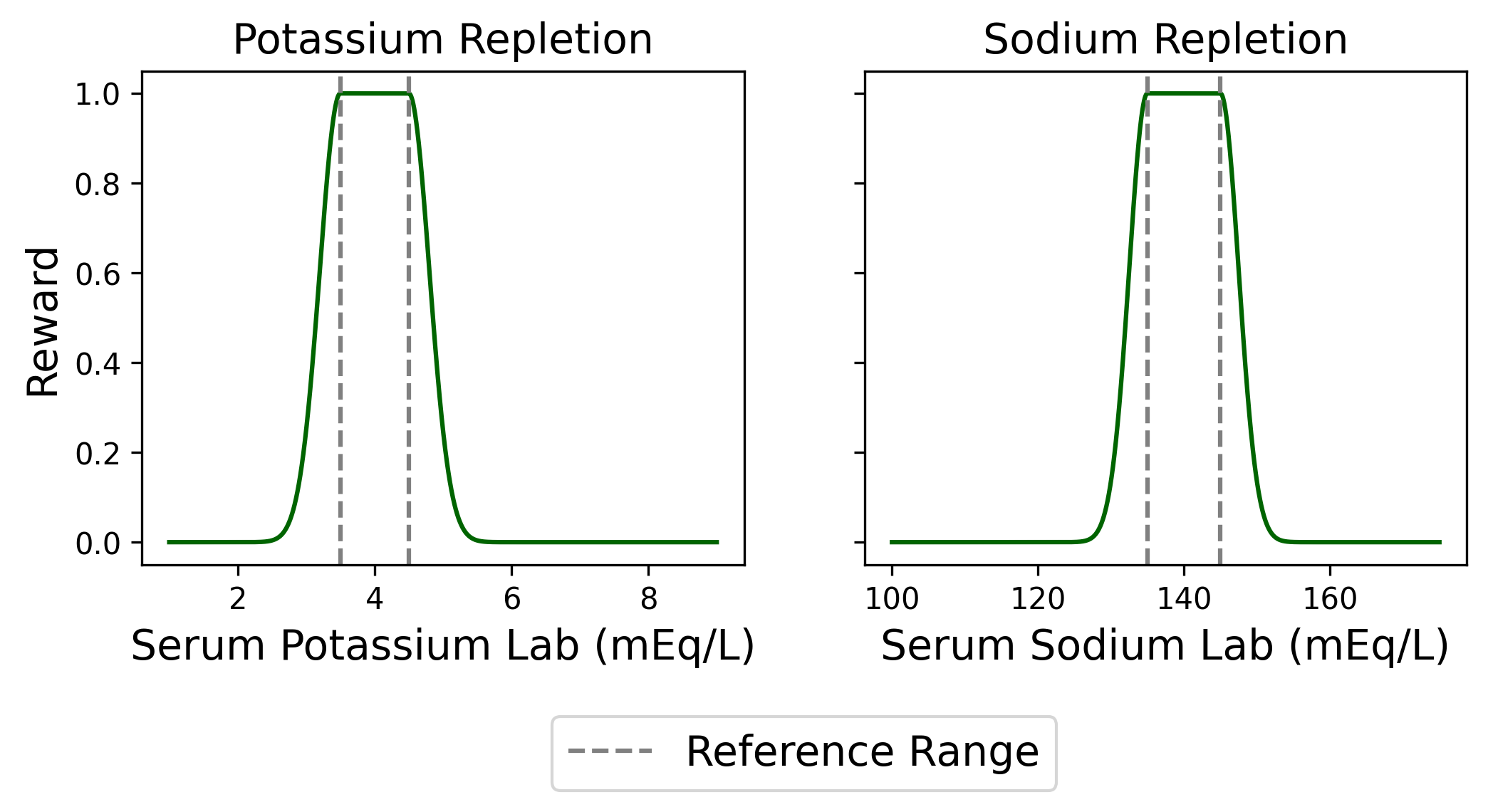}
    \caption{\textbf{Reward functions for both decision-making tasks are a function of the corresponding reference range. } Reward is bounded in the range $[0, 1]$, attaining its maximum when the lab value falls within the corresponding clinical reference range ($3.5-4.5$ mEq/L for serum potassium, and $135-145$ mEq/L for serum sodium). As the lab value deviates from this range, the reward decreases according to a Gaussian decay, with the lowest rewards assigned to critically low or high values. }
    \label{fig:rewards}
\end{figure}

We represent each patient context using the following 15 features: age, gender, weight, height, heart rate, respiratory rate, oxygen saturation pulseoxymetry, systolic blood pressure, diastolic blood pressure, serum creatinine lab, administered NaCl 0.9\%, administered dextrose 5\%, administered propofol, administered norepinephrine, and administered insulin. We choose these features due to their relevance in being able to predict downstream serum potassium and serum sodium labs~\citep{uptodate}. The reward function for both tasks is visualized in \Cref{fig:rewards}. 

When we report the accuracy of the LLM in predicting downstream lab values, we use weighted F1 score. The classes of predictions for potassium (all in mEq/L) are [$<3.2$, $>= 3.2$ and $< 5, >= 5$ and $< 6, >= 6$]. The classes of predictions for sodium (all in mEq/L) are [$<118$, $>=118$ and $< 135$, $>=135$ and $<152$, $>=152$ and $<169$, $>=169$]. 

\section{Prompts}
\label{apd:prompts}
\begin{figure*}
    \centering
    \includegraphics[width=\linewidth]{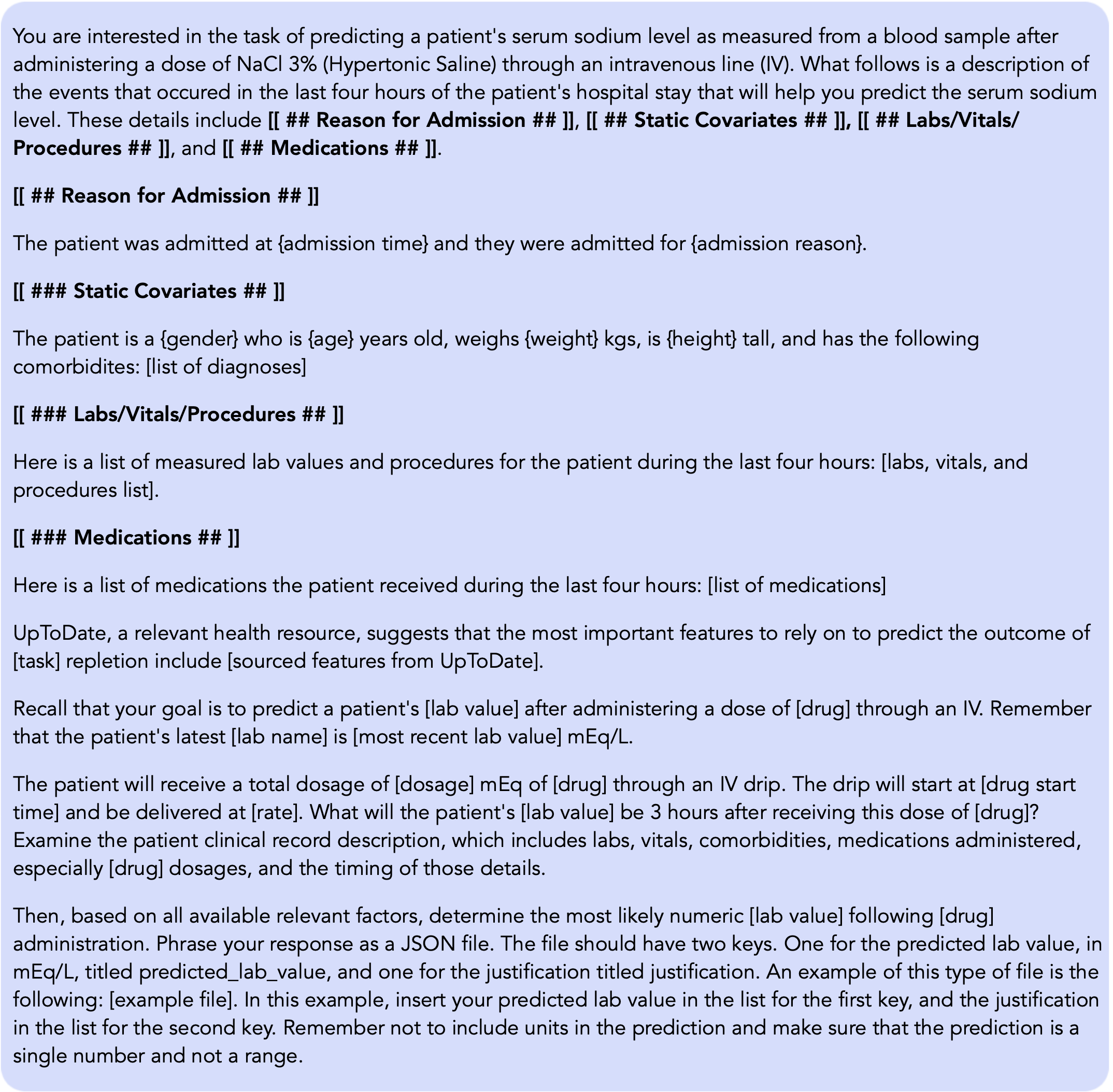}
    \caption{\textbf{LLMs can be prompted to construct downstream lab value predictions.} The prompt contains separate components that first describe the patient's clinical state four hours prior to receiving treatment, and then contains instructions to perform the lab value prediction. The prompt includes relevant information from UpToDate, a clinical resource, to help an LLM identify which features in the medical record are most predictive.}
    \label{fig:prompt_example}
\end{figure*}
Here we include the format of the prompt used to query downstream lab value predictions. The format is consistent across both the potassium and sodium repletion tasks, and varies only based on individual patient details. The prompt consists of five components: task information, static covariates, labs and medicines, domain information from UpToDate, and a prediction query. An example prompt is shown in \Cref{fig:prompt_example}. 

\section{Additional Empirical Results}
\label{apd:empirical_results}
\begin{figure*}[htbp]
\floatconts
  {fig:lab_preds}
  {\caption{\textbf{LLMs can accurately predict sodium and potassium lab values in MIMIC-IV.} 
  Predictions are evaluated using weighted F1 scores across clinically relevant lab value categories. 
  The black line denotes perfect agreement with ground truth, and predictions from a different LLM are reported in different colors.}}
  {%
    \subfigure[Potassium lab predictions.]{\label{fig:potassium_pred}%
      \includegraphics[width=0.8\linewidth]{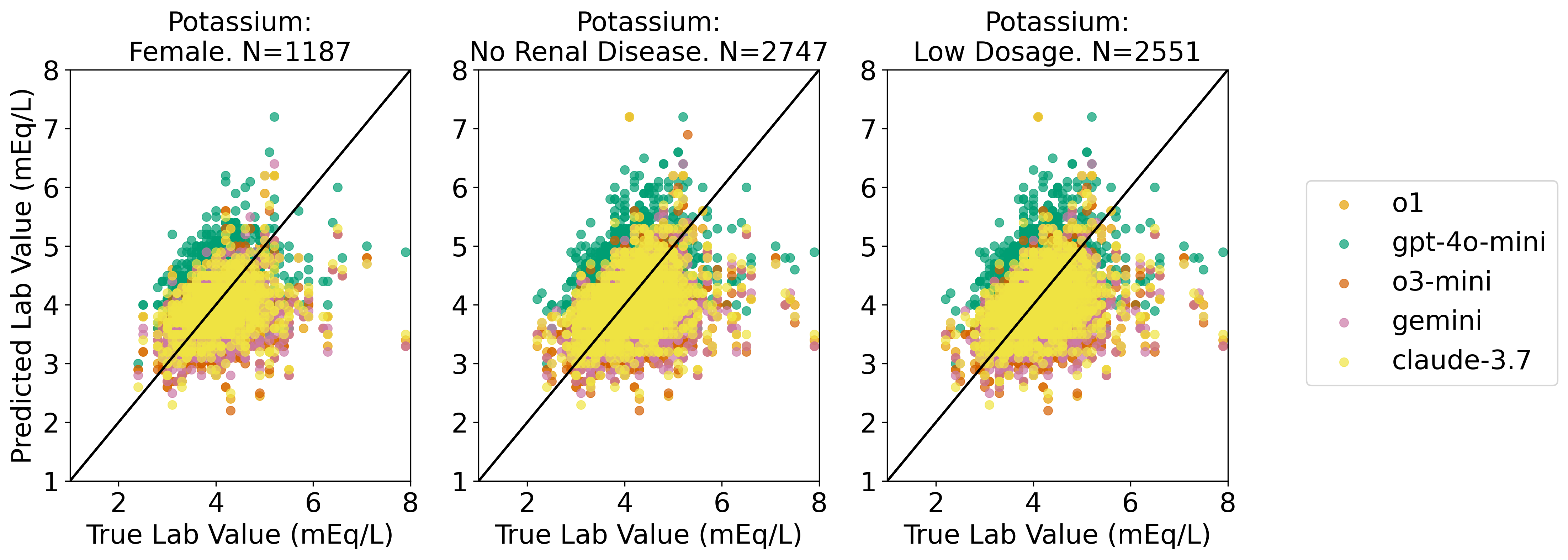}}%
    \qquad
    \subfigure[Sodium lab predictions.]{\label{fig:sodium_pred}%
      \includegraphics[width=0.8\linewidth]{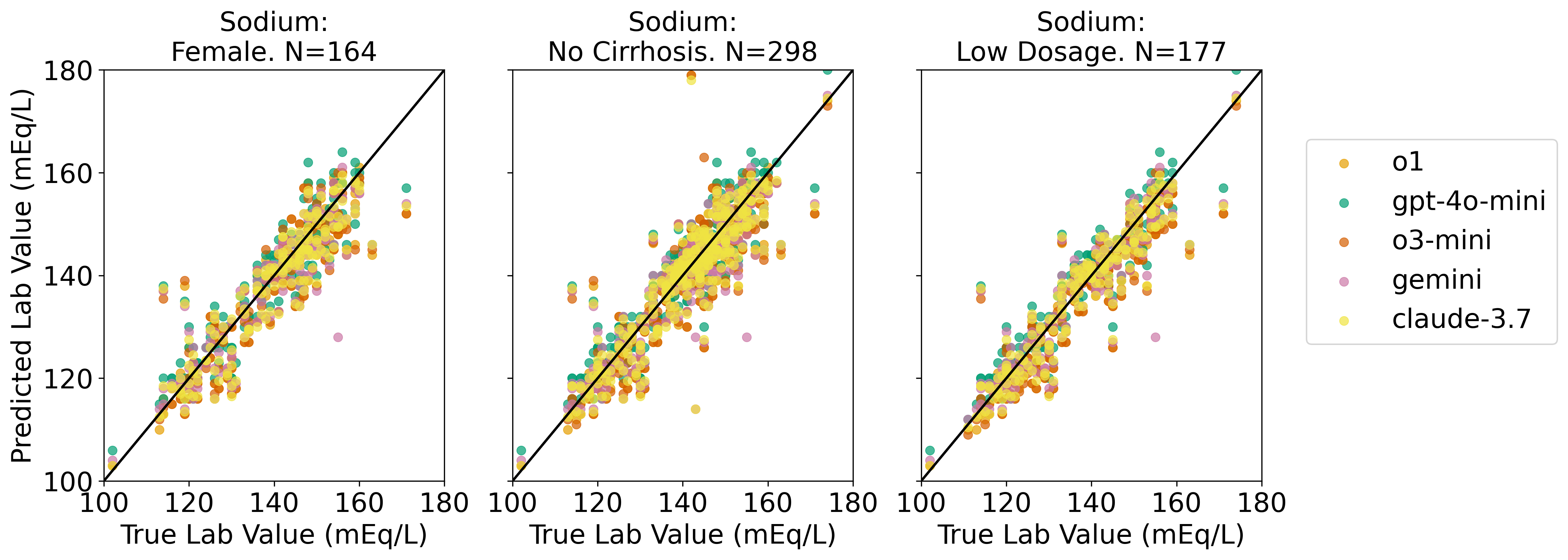}}
  }
\end{figure*}
\begin{figure*}[htbp]
\floatconts
  {fig:ope_hn}
  {\caption{\textbf{LLM-generated counterfactual annotations improve OPE estimates for sodium repletion in settings with high divergence between behavior and target policies.} We report results using the $DM$ baseline (blue) which uses no annotations, and $DM^+$ with annotations from different LLMs in other colors. Error bars represent standard error across $500$ bootstrapped datasets, truncated at 0 when necessary.}}
{\includegraphics[width=0.9\linewidth]{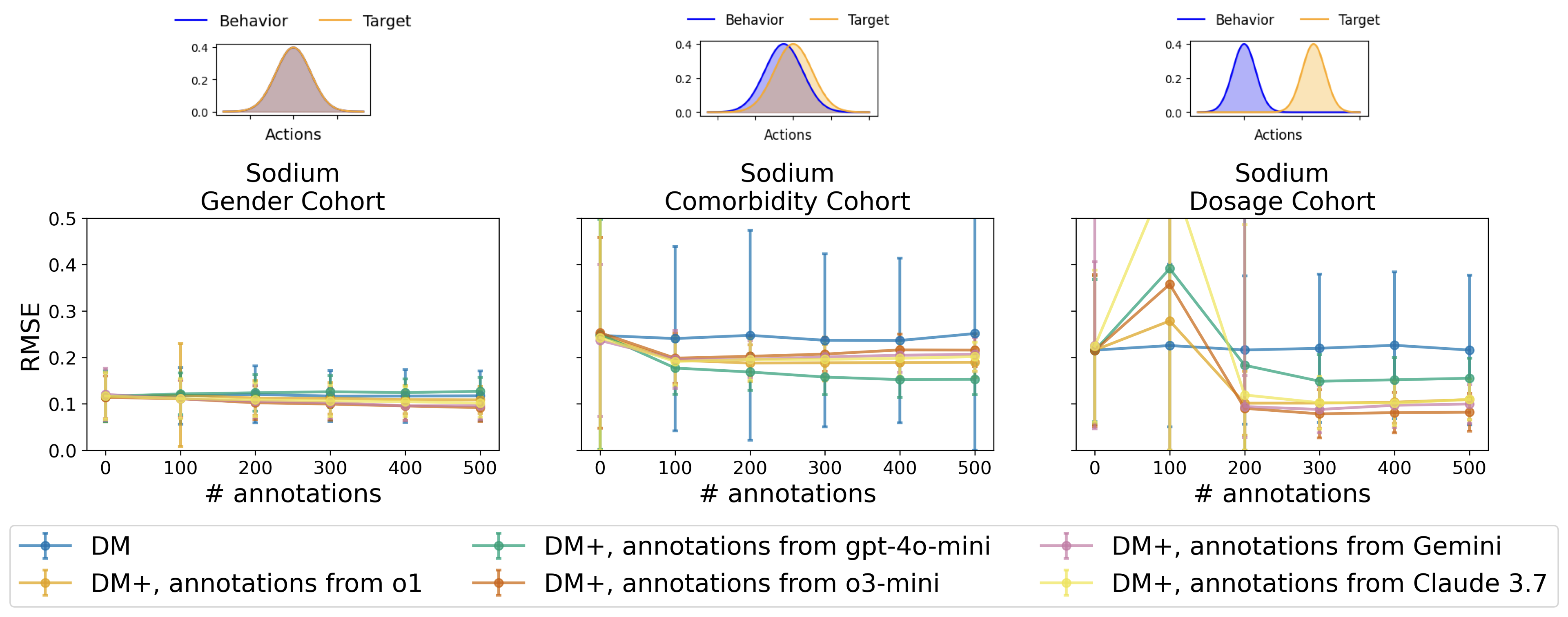}}
\end{figure*}
Here, we include additional empirical results to support our claims in the main text. First, we report figures that demonstrate the quality of downstream lab predictions across LLMs for both potassium and sodium lab predictions (\Cref{fig:lab_preds}). Our results conclude similarly to those reported in \Cref{tab:lab_preds}, suggesting that LLMs can predict downstream lab values within clinically relevant degrees of error. 

Furthermore, we investigate whether the age and gender of the patient affects the accuracy of the LLM in predicting potassium and sodium levels. We find that the prediction error varies substantially depending on the model and trends are not consistent given a patient's age or gender. (\Cref{fig:apd_errors}). 

\begin{figure}[htbp]
\floatconts
  {fig:apd_errors}
  {\caption{\textbf{Model prediction error varies depending on the patient's age and gender.} However, the trends are not consistent as observed for both sodium prediction error (\Cref{fig:subfig:sodium}) and potassium prediction error (\Cref{fig:subfig:potassium}).}}
  {%
    \subfigure[Potassium Lab Prediction Errors]{\label{fig:subfig:potassium}%
      \includegraphics[width=0.45\linewidth]{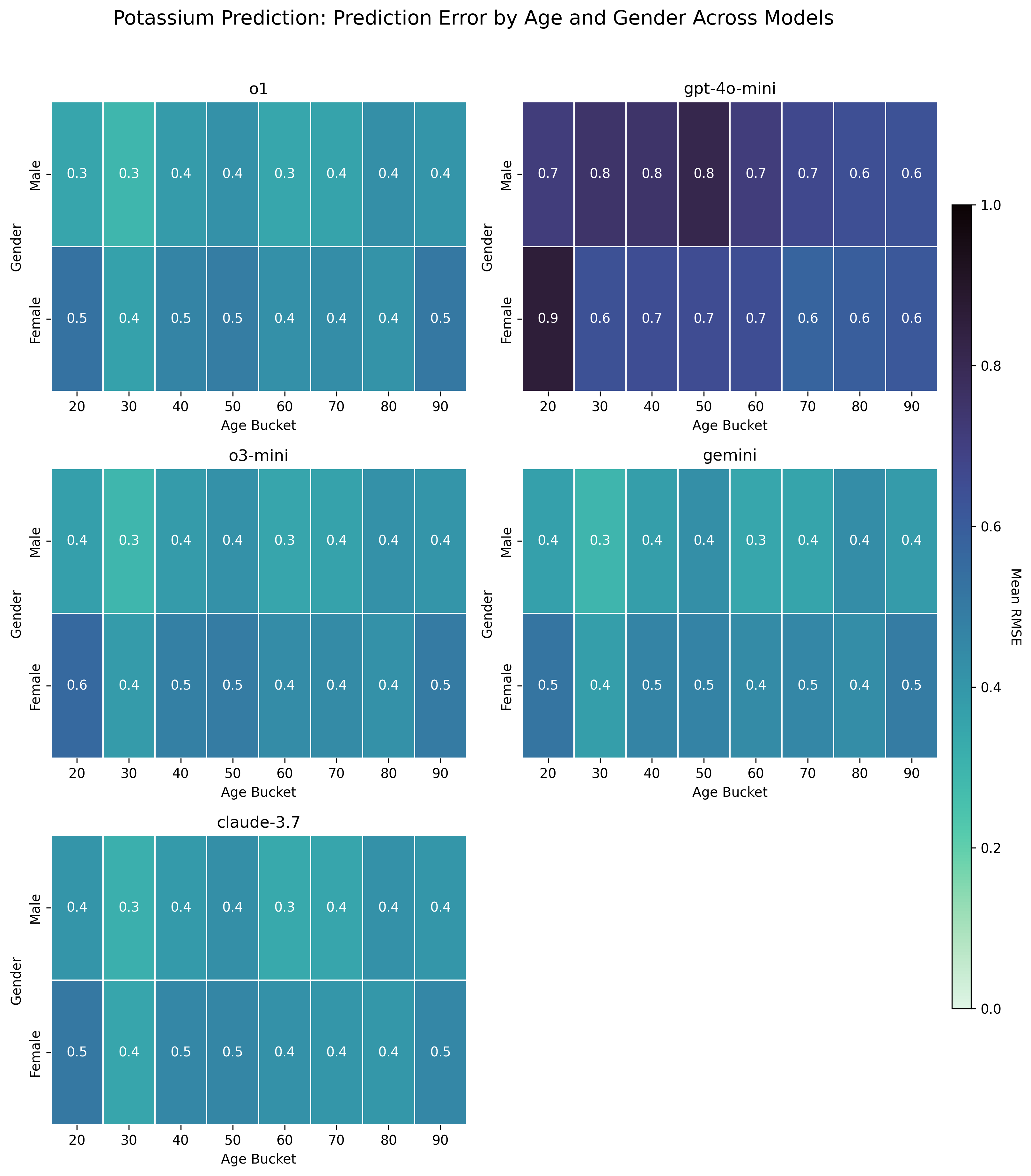}}%
    \qquad
    \subfigure[Sodium Lab Prediction Errors]{\label{fig:subfig:sodium}%
      \includegraphics[width=0.45\linewidth]{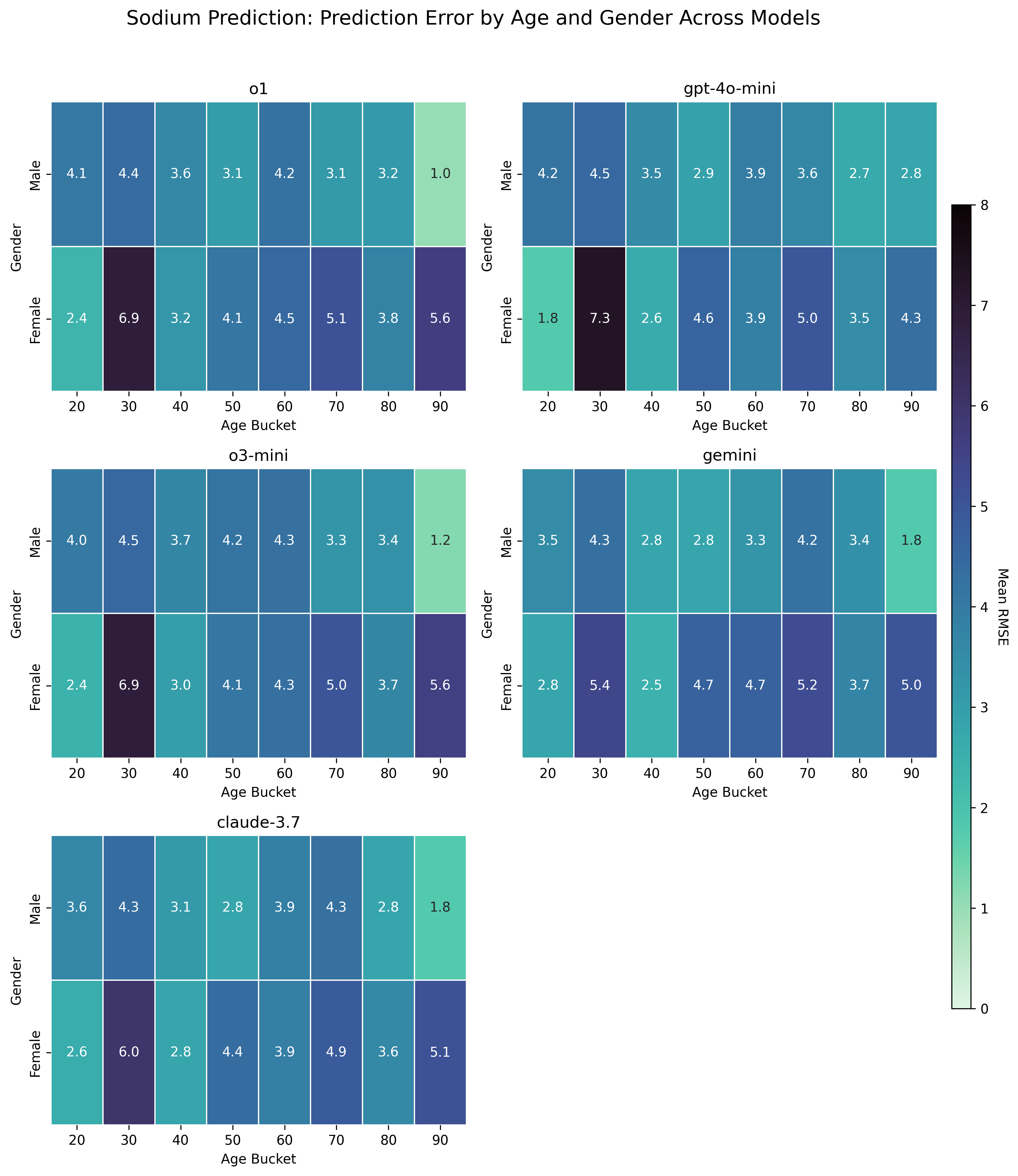}}
  }
\end{figure}

Now we discuss the utility of LLM-generated counterfactual annotations within the sodium repletion task (\Cref{fig:ope_hn}). Similar to the potassium repletion results, we find that LLM-generated counterfactual annotations help most when there is substantial divergence between the actions observed in the behavior and target policies. Just as in the potassium task results, the most improvement due to annotations occurs in the sodium dosage cohort. 

Finally, we report entropy and further annotations results for the sodium repletion task, suggesting that, similar to the potassium repletion task, that more annotations may yield only marginal gains (\Cref{fig:practical_hn}). 
\begin{figure}[htbp]
\floatconts
  {fig:practical_hn}
  {\caption{\textbf{Combining annotation sources yields limited returns in the sodium repletion task.} (Top) We compare $DM^+$ with the two best-performing LLMs for sodium repletion (yellow, green) and two aggregation methods: pooling predictions (orange) and averaging annotations (pink). Error bars show standard error over $500$ bootstrapped datasets, truncated at 0. (Bottom) Marginal entropy over the action space $H(A)$ when adding counterfactual annotations to the behavior dataset for the sodium cohort. The horizontal dashed line marks the maximum possible entropy.
  }}
{\includegraphics[width=0.5\linewidth]{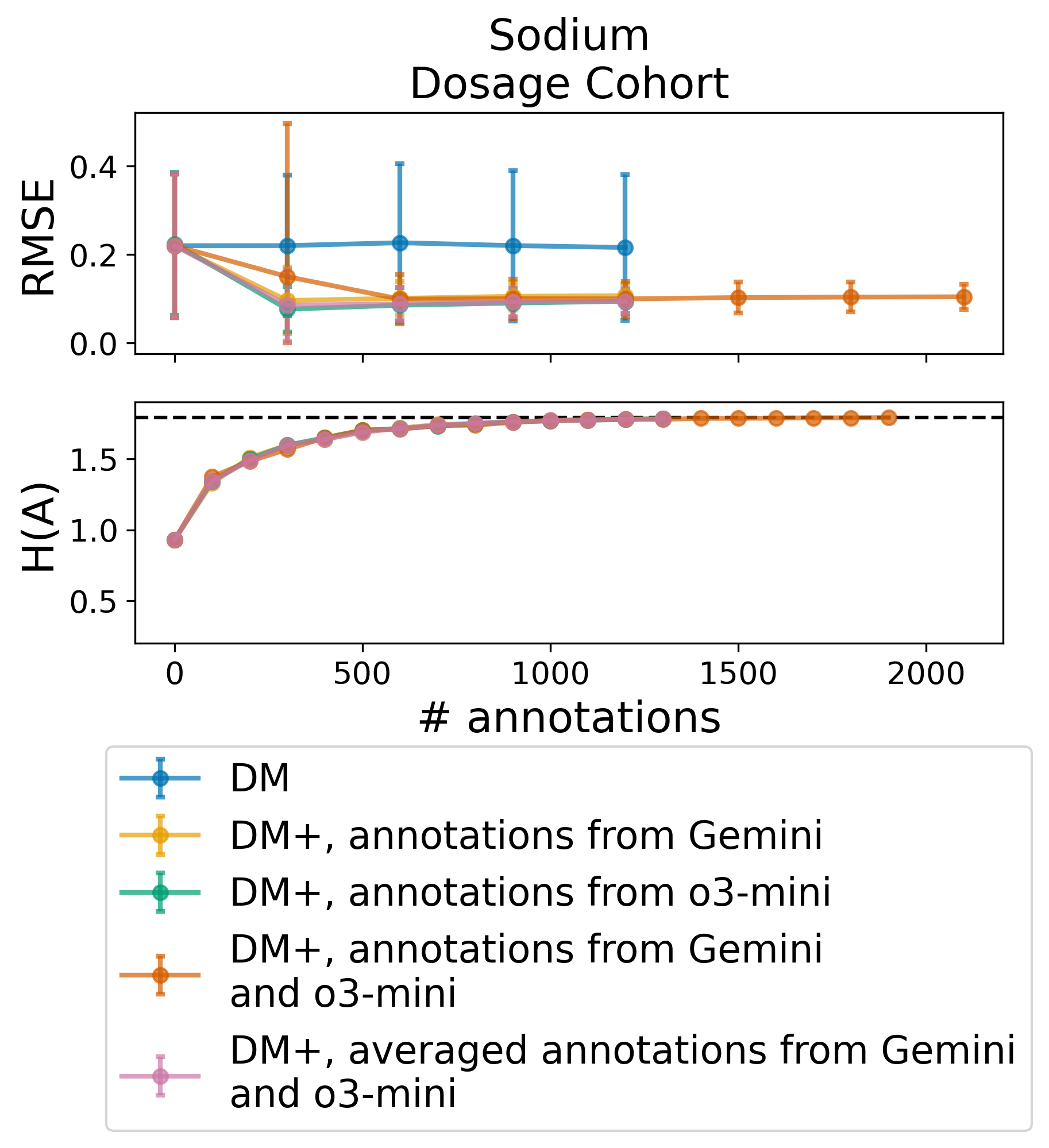}}
\end{figure}

\end{document}